\documentclass[11pt]{article}
	
\newcommand{\blind}{0}

\makeatletter
\renewcommand\section{\@startsection {section}{1}{\z@}%
                                   {-3.5ex \@plus -1ex \@minus -.2ex}%
                                   {2.3ex \@plus.2ex}%
                                   {\normalfont\fontfamily{phv}\fontsize{16}{19}\bfseries}}
\renewcommand\subsection{\@startsection{subsection}{2}{\z@}%
                                     {-3.25ex\@plus -1ex \@minus -.2ex}%
                                     {1.5ex \@plus .2ex}%
                                     {\normalfont\fontfamily{phv}\fontsize{14}{17}\bfseries}}
\renewcommand\subsubsection{\@startsection{subsubsection}{3}{\z@}%
                                    {-3.25ex\@plus -1ex \@minus -.2ex}%
                                     {1.5ex \@plus .2ex}%
                                     {\normalfont\normalsize\fontfamily{phv}\fontsize{14}{17}\selectfont}}
\makeatother
	
\usepackage{amsmath}
\usepackage{graphicx}
\usepackage{enumerate}
\usepackage{xcolor}
\usepackage{natbib} 
\usepackage{url} 

\usepackage{amsfonts, amsthm, latexsym, amssymb}
\usepackage{hyperref}

\usepackage{algorithm}
\usepackage{algpseudocode}

\DeclareMathOperator*{\argmax}{argmax}

\begin{document}
		
\def\spacingset#1{\renewcommand{\baselinestretch}%
    {#1}\small\normalsize} \spacingset{1}
    
    \if0\blind
    {
        \title{\bf Trust Region Constrained Bayesian Optimization with Penalized Constraint Handling}
        \author{Raju Chowdhury*$^{1}$, Tanmay Sen$^{1}$, Biswabrata Pradhan$^{1}$\\
        $^{1}$Statistical Quality Control and Operations Research Unit, \\Indian Statistical Institute, Kolkata, India.\\
        \\\textit{*Corresponding author:} \texttt{raju.chowdhury06@gmail.com}}
        \date{}
        \maketitle
    } \fi
    
    \if1\blind
    {

        \title{\bf \emph{IISE Transactions} \LaTeX \ Template}
        \author{Author information is purposely removed for double-blind review}
			
\bigskip
        \bigskip
        \bigskip
        \begin{center}
            {\LARGE\bf High CBO}
        \end{center}
        \medskip
    } \fi
    \bigskip
    
\begin{abstract}
Constrained optimization in high-dimensional black-box settings is difficult due to expensive evaluations, the lack of gradient information, and complex feasibility regions. In this work, we propose a Bayesian optimization method that combines a penalty formulation, a surrogate model, and a trust region strategy. The constrained problem is converted to an unconstrained form by penalizing constraint violations, which provides a unified modeling framework. A trust region restricts the search to a local region around the current best solution, which improves stability and efficiency in high dimensions. Within this region, we use the Expected Improvement acquisition function to select evaluation points by balancing improvement and uncertainty.

The proposed Trust Region method integrates penalty-based constraint handling with local surrogate modeling. This combination enables efficient exploration of feasible regions while maintaining sample efficiency. We compare the proposed method with state-of-the-art methods on synthetic and real-world high-dimensional constrained optimization problems. The results show that the method identifies high-quality feasible solutions with fewer evaluations and maintains stable performance across different settings.
\end{abstract}

\noindent%

\newpage
\spacingset{1.5} 

\allowdisplaybreaks
\section{Introduction} \label{sec1}
In many real-world settings, optimization problems arise without an explicit mathematical form. These problems are treated as black-box optimization problems, where the objective and constraint functions can only be evaluated at selected input points. Such problems appear in engineering design, model tuning, and scientific simulations.
The difficulty becomes more pronounced in high-dimensional spaces due to the curse of dimensionality \citep{powell2019}. As the number of variables grows, the search space becomes large and harder to explore with limited evaluations. It becomes difficult to efficiently identify promising regions. The presence of constraints further reduces the feasible region, which is often small and irregular, making the search for feasible and optimal solutions more challenging.

Bayesian optimization (BO) \citep{,shahriari2016,gramacy2020,garnett2023} provides a principled approach for solving black-box optimization problems when function evaluations are expensive. It builds a probabilistic surrogate model of the unknown objective, often using Gaussian Processes, which gives both a prediction of the function value and a measure of uncertainty in unexplored regions. An acquisition function then selects new evaluation points by balancing improvement in the objective with uncertainty in the predictions. The method proceeds sequentially, with each new observation updating the surrogate model and improving future decisions. This approach is sample efficient and is well-suited for high-dimensional, and constrained settings where evaluations are costly, and the feasible region may be limited.

However, in high-dimensional settings \citep{greenhill2020,papenmeier2026}, standard BO often performs poorly because surrogate models become less accurate over the full domain, and optimizing the acquisition function becomes difficult. This leads to inefficient exploration and unstable search behavior. As the dimension increases, the search space grows rapidly, making it harder to identify promising regions with limited evaluations. The presence of constraints further increases the difficulty \citep{amini2025}, as the feasible region may be small, irregular, and hard to locate. Together, these challenges make it difficult to maintain reliable performance and efficiently identify feasible and optimal solutions.

The classical optimization literature has long studied methods for handling constraints, including penalty functions, augmented Lagrangian methods, and barrier functions \citep{nocedal2006}. These approaches transform constrained problems into unconstrained ones or decompose them into simpler subproblems that can be solved iteratively.
Building on these ideas, \citet{gramacy2016} combined classical optimization with BO and proposed an Augmented Lagrangian-based BO framework, where the acquisition function is defined on the predictive mean surface. Inspired by this approach, \citet{ariafar2019} introduced a framework, ADMMBO, that uses the big-\(M\) penalty and the Alternating Directions Method of Multipliers together with the Expected Improvement acquisition function. In a similar direction, \citet{pourmohamad2022} developed a log-barrier based BO framework and constructed the acquisition function using the predictive mean. Recently, \citet{upadhye2024} proposed a big-\(M\) penalty-based approach, known as \(M\)-LCBBO, in which a constrained version of the Lower Confidence Bound acquisition function is derived.

High-dimensional constrained optimization is challenging due to the large search space and difficulty in identifying feasible regions. Global surrogate models often become unreliable, which motivates local modeling strategies. Trust region based methods restrict the search to local regions and adapt their size based on progress. TuRBO \citep{eriksson2019} demonstrates improved performance in high dimensions, while SCBO \citep{eriksson2021} extends this idea to constrained settings, defines a hyperrectangular trust region. FuRBO \citep{paolo2025} further modifies the trust region using a hyperspherical domain to improve exploration. These approaches demonstrate that local trust-region strategies are effective for high-dimensional constrained optimization.

We propose a trust region based BO framework, referred to as TR-\(M\)EI, to address high-dimentional constrained optimization problem. The framework integrates big-\(M\) penalty approach with surrogate modeling and a trust region strategy. In this setting, we use a modified Expected Improvement acquisition function defined over the penalized objective, balancing exploration and exploitation within the local trust region.

The article is organized as follows. Section \ref{sec2} presents the relevant background and existing trust region based BO methods. Section \ref{sec3} describes the proposed methodology along with the trust region strategy. Section \ref{sec4} presents the empirical comparison with existing methods on synthetic and real-world problems. Section \ref{sec5} concludes the article with a discussion and directions for future research.

\section{Background} \label{sec2}
In this Section, we briefly describe Bayesian Optimization and its main components, including surrogate modeling and Expected Improvement acquisition function for selecting evaluation points. We also outline a trust region based Bayesian optimization approach for high-dimensional settings.

\subsection{Bayesian Optimization} \label{ssec2.1}
Bayesian Optimization (BO) \citep{shahriari2016,frazier2018} is an effective method for optimizing black box functions. It is a derivative-free and sequential approach that aims to find the optimum with a small number of evaluations. BO has two main components. The first is a surrogate model that approximates the original function. The second is an acquisition function that guides the selection of the next sample point. A widely used surrogate is the Gaussian Process (GP) \citep{Rasmussen2006}, defined by its mean and covariance or kernel function. Among acquisition functions, Expected Improvement (EI) \citep{garnett2023} is well structured, easy to compute, and widely used. Below, we give a brief mathematical description of GP surrogate modelling and EI.

\noindent
\textbf{Surrogate Modelling:}\\
Let \(\mathcal{D}_n = \{x_i, f(x_i)\}_{i=1}^{n}\) denote the observed input output pairs. We model these data using a GP \(Y(\cdot)\) with mean function \(m(\cdot)\) and covariance function \(k(\cdot,\cdot)\). The model is trained on \(m(\cdot)\) and \(k(\cdot,\cdot)\), estimated using \(\mathcal{D}_n\).
Next, for a new input \(x\), the trained model \(Y(\cdot)\) gives a posterior prediction that follows a normal distribution with mean \(\mu(x)\) and variance \(\sigma^2(x)\) \citep{Rasmussen2006}, that is
\begin{align*}
Y(x) \mid \mathcal{D}_n &\sim \mathcal{N}\big(\mu(x), \sigma^2(x)\big)\\
\mu(x) &= k(x, X)\, K^{-1} \mathbf{y}\\
\sigma^2(x) &= k(x,x) - k(x, X)\, K^{-1} k(X, x)
\end{align*}
where \(X = [x_1, \dots, x_n]\), \(\mathbf{y} = [f(x_1), \dots, f(x_n)]^\top\), and \(K\) is the covariance matrix with entries \(K_{ij} = k(x_i, x_j)\). Thus, these posterior predictive quantities, \(\mu(x)\) and \(\sigma^2(x)\), are used in the EI acquisition function.

\newpage
\noindent
\textbf{Expected Improvement (EI):}\\
The idea of EI is based on the improvement statistic, defined as
\[
\mathcal{I}(x) = \max\big(0, f^* - Y(x)\big),
\]
which measures the improvement over the current best value \(f^* = \min(f(x))\). Based on this, \cite{jones1998} defines the EI acquisition function as the expectation of \(\mathcal{I}(x)\). The expression of EI is
\begin{equation}
\label{eq2.1}
\text{EI}(x) = \big(f^* - \mu(x)\big)\Phi\Bigg(\frac{f^* - \mu(x)}{\sigma(x)}\Bigg) + \sigma(x)\phi\Bigg(\frac{f^* - \mu(x)}{\sigma(x)}\Bigg),
\end{equation}
where \(\mu(x)\) and \(\sigma(x)\) are the posterior predictive mean and standard deviation of \(Y(x)\) at some point \(x\), and \(\Phi(\cdot)\) and \(\phi(\cdot)\) denote the cumulative distribution function and probability density function of the standard normal distribution.
EI balances the trade-off between exploration and exploitation during the search process. The first term emphasizes exploitation by favoring points with high predicted values. The second term emphasizes exploration by favoring points with high uncertainty.
The next evaluation point is obtained by maximizing EI, that is,
\(x^* = \argmax \text{EI}(x),\)
which is then used for the next function evaluation and model update.

\subsection{Trust Region CBO Approaches} \label{ssec2.2}
Solving a constrained optimization problem in high-dimensional space using the BO algorithm is challenging due to the large search space and the influence of constraints. To handle high dimensionality, the trust region method restricts the search to a local region around the current best solution. This region adapts during the optimization process; it expands when improvements are observed and shrinks when progress slows. This local search improves efficiency and stability in high-dimensional settings. Here, we discuss two recently proposed trust-region based BO approaches.

\subsubsection{Scalable
Constrained Bayesian Optimization (SCBO)}
A framework based on the trust-region algorithm \citep{yuan2015}, \cite{eriksson2019} proposed the trust region BO (TuRBO) algorithm for solving the unconstrained black-box optimization problem in a high-dimensional setting. Later, \cite{eriksson2021} extends TuRBO to Scalable Constrained Bayesian Optimization (SCBO) to handle black-box high-dimensional optimization problems with constraints. SCBO initialized a trust region, that is, a hyperrectangle, around the best feasible solution found so far. If no feasible point is available, it is centered at the point with the smallest constraint violation. 

At each iteration, SCBO generates a candidate point within the current trust region using Thompson sampling (TS) \citep{thompson1933} for the next function evaluation. TS balance improvement from the current best value and the feasibility of constraints. Also, the size of the trust region is adjusted during the optimization process. It expands when successful improvements are observed and shrinks when the search fails to make progress. This adaptive strategy helps maintain efficient, stable performance in high-dimensional, constrained settings.

\subsubsection{Feasibility-Driven Trust Region Bayesian Optimization (FuRBO)}
The FuRBO framework \citep{paolo2025} builds on the TuRBO algorithm \citep{eriksson2019}, but differs in the way the trust region is defined. Similar to SCBO, the trust region is centered at the best feasible point, or at the point with the smallest constraint violation when no feasible solution is available. However, FuRBO defines the trust region as a hypersphere instead of a hyperrectangle.

This spherical region provides an isotropic search space around the center, which avoids bias toward coordinate directions. Candidate points are sampled within this hypersphere, which leads to a more uniform exploration of the local region. The size of the trust region is adapted during the optimization process. It expands when improvements are observed and shrinks when progress slows down. This design improves the robustness of the search in high-dimensional constrained settings.

\section{Methodology} \label{sec3}
This Section discusses the methodology that combines the big-$M$ penalty approach for constraint handling with the trust region to address the high dimensionality of the optimization problem.
\subsection{Formulation}
The general constrained optimization problem is defined as
\begin{equation}
\label{eq3.1}
\begin{aligned}
    \min_{\mathbf{x}\in \mathcal{B}}~ &f(\mathbf{x}) \\
    \text{subject to,}~ &g_j(\mathbf{x}) \leq 0, ~ j=1,2,\dots,J.
\end{aligned}
\end{equation}
where \(\mathcal{B}\) is a high-dimensional search space, typically a bounded domain in \(\mathbb{R}^D\). In this formulation, \(f(\mathbf{x})\) denotes the objective function, and \(g_j(\mathbf{x})\) represents the set of constraint functions.

All functions in \eqref{eq3.1} are treated as black-box. Their analytical forms are unknown, and gradient information is not available. Only point-wise evaluations of \(f(\mathbf{x})\) and \(g_j(\mathbf{x})\) can be obtained, which are often expensive to compute. The goal is to identify a feasible solution that satisfies all constraints while minimizing the objective function using a limited number of evaluations.

In recent years, converting a constrained problem into an unconstrained one has become common in the BO literature \citep{gramacy2016,ariafar2019,upadhye2024,zhao2024,chowdhury2025}. These approaches rely on penalty-based methods from classical optimization \citep{nocedal2006}, where constraint violations are incorporated into the objective function.

In this article, we adopt the big-\(M\) penalty approach, which imposes a penalty when constraints are violated. The constrained problem is reformulated as the following unconstrained problem \citep{ariafar2019,upadhye2024,chowdhury2025},
\begin{equation}
\label{eq3.2}
    \min_{\mathbf{x}\in \mathcal{B}}~ F(\mathbf{x}) = f(\mathbf{x}) + M \sum_{j=1}^{J}\mathbb{I}\big(g_j(\mathbf{x}) > 0 \big),
\end{equation}
where \(\mathbb{I}(\cdot)\) is the indicator function that returns \(1\) if the constraint is violated and \(0\) otherwise, and \(M\) is a large positive constant that penalizes infeasible solutions.

This formulation assigns an additional cost to any point that violates one or more constraints. As a result, feasible solutions are always preferred over infeasible ones when \(M\) is sufficiently large. The transformed objective \(F(\mathbf{x})\) can then be optimized using standard BO methods without explicitly handling constraints.

Now, we model the unconstrained objective in \eqref{eq3.2} using independent GPs for the objective and each constraint. Let \(Y_f(\mathbf{x})\) denote the GP model for the objective and \(Y_{g_j}(\mathbf{x})\) denote the GP model for the \(j\)-th constraint. The resulting surrogate for the penalized objective is given by
\begin{equation}
\label{eq3.3}
    Y_F(\mathbf{x}) = Y_f(\mathbf{x}) + M \sum_{j=1}^{J}\mathbb{I}\big(Y_{g_j}(\mathbf{x}) > 0 \big).
\end{equation}

This formulation combines the predictive distributions of the objective and constraints. Since each component is modeled independently, we can derive the predictive mean and variance of \(Y_F(\mathbf{x})\) at a new point \(x\) as
\begin{align}
    \mathbb{E}\big[Y_F(x)\big] &= \mu_F(x) = \mu_f(x) + M \sum_{j=1}^{J} p\big(Y_{g_j}(x) > 0 \big), \label{eq3.4}\\
    \mathbb{V}\big[Y_F(x)\big] &= \sigma^2_F(x) = \sigma^2_f(x) + M^2 \sum_{j=1}^{J} p\big(Y_{g_j}(x) > 0 \big)\Big(1 - p\big(Y_{g_j}(x) > 0 \big)\Big), \label{eq3.5}
\end{align}
where \(p\big(Y_{g_j}(x) > 0 \big) = 1 - \Phi\bigg(-\frac{\mu_{g_j}(x)}{\sigma_{g_j}(x)}\bigg)\) denotes the probability of constraint violation.

The mean \(\mu_F(x)\) increases with the probability of infeasibility, which penalizes regions that are likely to violate constraints. The variance \(\sigma^2_F(x)\) reflects both the uncertainty in the objective and the uncertainty in constraint satisfaction.

Using these quantities, we define the penalized EI acquisition function \citep{chowdhury2025} as
\begin{equation}
\label{eq3.6}
\text{$M$EI}(x) = \big(F^* - \mu_F(x)\big)\Phi\Bigg(\frac{F^* - \mu_F(x)}{\sigma_F(x)}\Bigg) + \sigma_F(x)\phi\Bigg(\frac{F^* - \mu_F(x)}{\sigma_F(x)}\Bigg),
\end{equation}
where \(F^* = \min\big(F(x)\big)\). This acquisition function guides the search toward regions with low objective values while accounting for the likelihood of constraint satisfaction.

Next, we integrate the $M$EI acquisition function with a trust region strategy to obtain Trust Region $M$EI (TR-\(M\)EI). Instead of optimizing $M$EI over the entire search space \(\mathcal{B}\), the optimization is restricted to a local trust region centered at the current best solution.

\subsection{Trust-Region \texorpdfstring{$M$EI}{}}
In a high-dimensional constrained optimization setting, the trust region approach is effective for locating feasible and optimal solutions. In the literature, SCBO and FuRBO are two trust region based methods that differ in how the next sample point is selected during the BO procedure. As discussed in Section \ref{ssec2.2}, SCBO defines the trust region over a hyperrectangle domain, while FuRBO defines it over a hypersphere domain.

In this article, we embrace the hyperrectangle domain setting and use coordinate-wise perturbation to generate candidate points within the trust region. We also employ the EI acquisition function within this local region.

At each iteration, a trust region is defined around the best observed point. Candidate points are generated within this region and evaluated using the \(M\)EI criterion. The size of the trust region is updated during the optimization process. It expands when improvements are observed and shrinks when no progress is made. This adaptive mechanism focuses the search in promising regions while maintaining stability.

By combining the big-\(M\) penalty-based formulation with a trust region strategy, TR-\(M\)EI improves efficiency in high-dimensional constrained problems. It reduces the effect of the large search space and guides the optimization toward feasible and optimal regions.

Further, we compare the proposed TR-\(M\)EI method with SCBO and FuRBO to highlight the differences:
\begin{itemize}
\item TR-\(M\)EI solves an unconstrained version of the constrained problem, while SCBO and FuRBO handle the constrained problem explicitly.
\item TR-\(M\)EI and SCBO define the hyperrectangular trust region, while FuRBO uses hyperspherical.
\item TR-\(M\)EI uses the standard EI acquisition function, while SCBO and FuRBO rely on Thompson sampling strategies.
\end{itemize}

\noindent
The advantages of the proposed methodology are as follows. First, the TR-\(M\)EI acquisition function admits an explicit mathematical expression. This allows direct computation and avoids the need for approximation or sampling based estimates. As a result, the method remains simple to implement and computationally efficient.

Second, the method can start from infeasible initial points. The big-\(M\) penalty formulation assigns higher objective values to constraint violations, which guides the search toward feasible regions even when no feasible solution is available at the beginning. This makes the approach robust in practical scenarios where feasible initialization is difficult.

Finally, the integration of the trust region strategy enables the method to handle high-dimensional constrained optimization problems effectively. By restricting the search to a local region and adapting its size based on observed progress, the method improves stability and reduces the effect of the large search space while maintaining convergence toward feasible and optimal solutions.

\section{Experiments} \label{sec4}
This section presents the comparison results of the proposed algorithm on a range of synthetic and real-world high-dimensional constrained optimization problems. The goal is to evaluate the method's performance in terms of solution quality, constraint satisfaction, and computational efficiency.

\begin{algorithm}[!ht]
\caption{Trust Region $M$-EI}
\label{algo:1}
\begin{algorithmic}[1]
\State Evaluate initial design $X$ and train surrogate models $(f, g_j)$
\State$x_{\text{best}} \leftarrow \arg\min_{x \in \mathcal{B}} S(X; f, g_j)$
\While{Optimization budget not exhausted}
    \State$TR \leftarrow \text{define\_TR}$
    \State$x_{\text{next}} \leftarrow \text{TR-$M$EI}((f, g_j), TR)$
    \State$f_{\text{next}} \leftarrow f(x_{\text{next}})$
    \State$g_j^{\text{next}} \leftarrow g_j(x_{\text{next}})$
    \State$S \leftarrow S \cup \{(x_{\text{next}}, f_{\text{next}}, g_j^{\text{next}})\}$
    \State Fit surrogate models over $S$
    \State Update $n_s$ and $n_f$
    \If{$n_s = \tau_s$ or $n_f = \tau_f$}
        \State$L \leftarrow \text{adjust}(L)$
    \EndIf
    \State $x_{\text{best}} \leftarrow \arg\min_{x \in \mathcal{X}} S(X; f, g_j)$
\EndWhile
\State\textbf{return} $x_{\text{best}}$

\end{algorithmic}
\end{algorithm}

\subsection{Experimental setup}
We evaluate the proposed TR-\(M\)EI algorithm against SCBO and FuRBO. The comparison focuses on their performance in high-dimensional constrained optimization problems, where both feasibility and sample efficiency are important.

For evaluation, we consider three representative test functions, namely Ackley, L$\hat{e}$vy, and Rastrigin. These functions capture different landscape characteristics such as separable, ill-conditioned, and multimodal behavior. Each function is defined in $20$ dimensions with moderate constraint difficulty. The same experimental setting is used for all methods to ensure a fair comparison. Each algorithm begins with an initial design of $2D$ points and is allowed a total evaluation budget of $50$ function evaluations. All experiments are repeated multiple times with different initial designs and random seeds to account for variability in performance.

For a fair comparison across all frameworks, the number of initial random points is kept the same. We set \(n = 2 \times d\), where \(d\) is the dimension of the problem. We use a zero mean function and a Mat$\acute{e}$rn \(5/2\) kernel for fitting the GP surrogate model. The results are averaged over $30$ runs and reported with one standard error to reflect variability. 

All implementations of TR-\(M\)EI, FuRBO, and SCBO are carried out in Python using the GPyTorch \citep{gardner2018} and BoTorch \citep{balandat2020} libraries. A common implementation framework is used for all methods to ensure consistency in modeling and evaluation. This setup allows a fair comparison across methods while maintaining computational efficiency. Further Algorithm \ref{algo:1} provides the flow-chart of the TR-$M$EI methodology.

Feasible solutions are always preferred over infeasible ones. Any infeasible point is assigned the worst objective value observed for the corresponding problem across all methods. This ensures that the evaluation properly reflects both optimality and constraint satisfaction.

\subsection{Results}
In this subsection, we compare TR-\(M\)EI, FuRBO, and SCBO on the Ackley, L$\hat{e}$vy, and Rastrigin synthetic problems in a $20$ dimensional constrained setting. These benchmark functions are widely used to evaluate optimization algorithms due to their complex landscapes. The Ackley function is characterized by a multimodal surface with many local minima, while the L$\hat{e}$vy function presents a challenging structure with strong nonlinearity. The mathematical formulations of the problems are given in \eqref{app_eq1}, \eqref{app_eq2}, \eqref{app_eq3}.
\begin{figure}[!ht]
    \centering
    \includegraphics[width=0.95\linewidth]{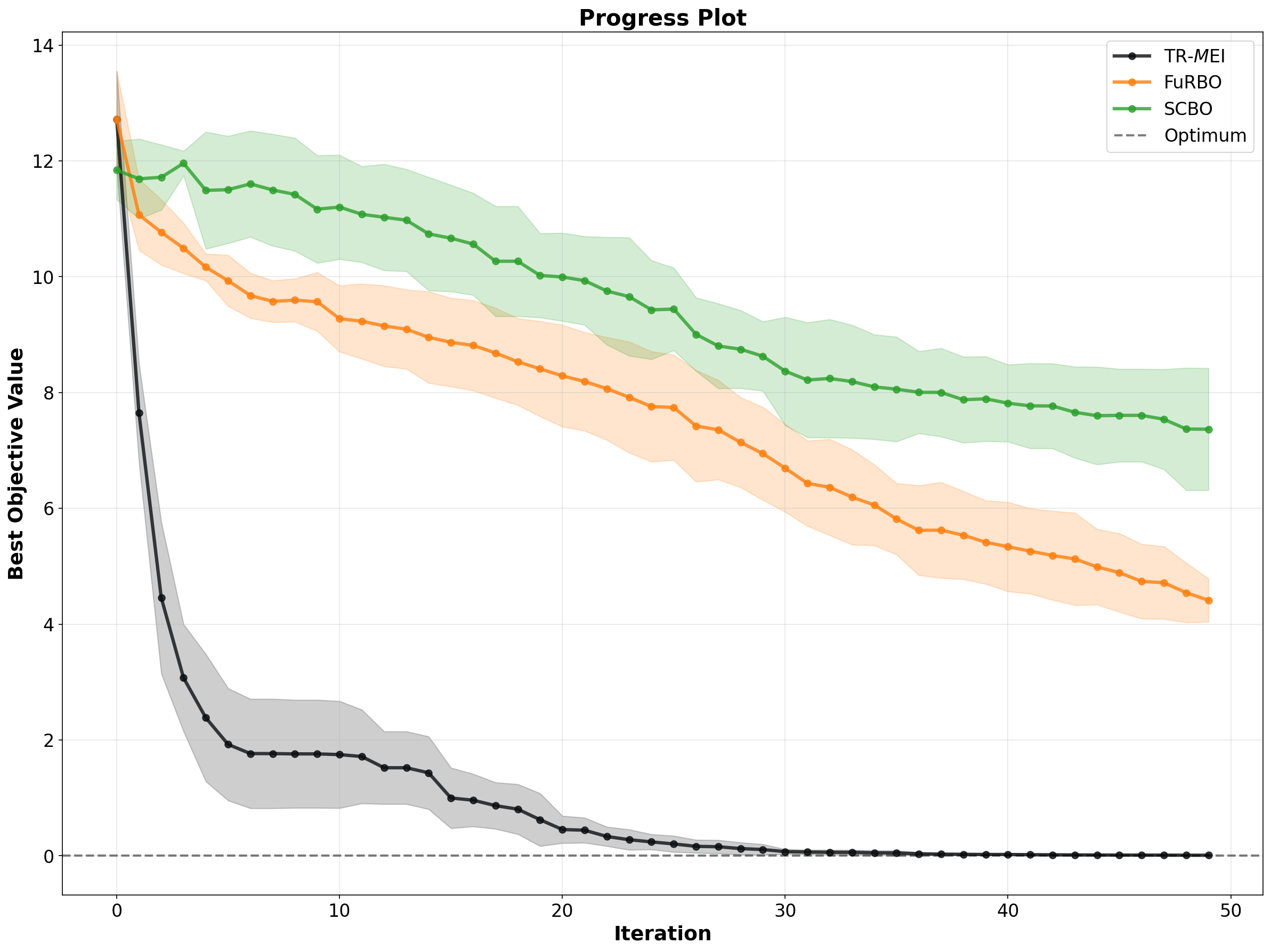}
    \caption{Average objective values obtained over $30$ repeated experiments on Ackley problem.}
    \label{fig:1}
\end{figure}

\begin{figure}[!ht]
    \centering
    \includegraphics[width=0.95\linewidth]{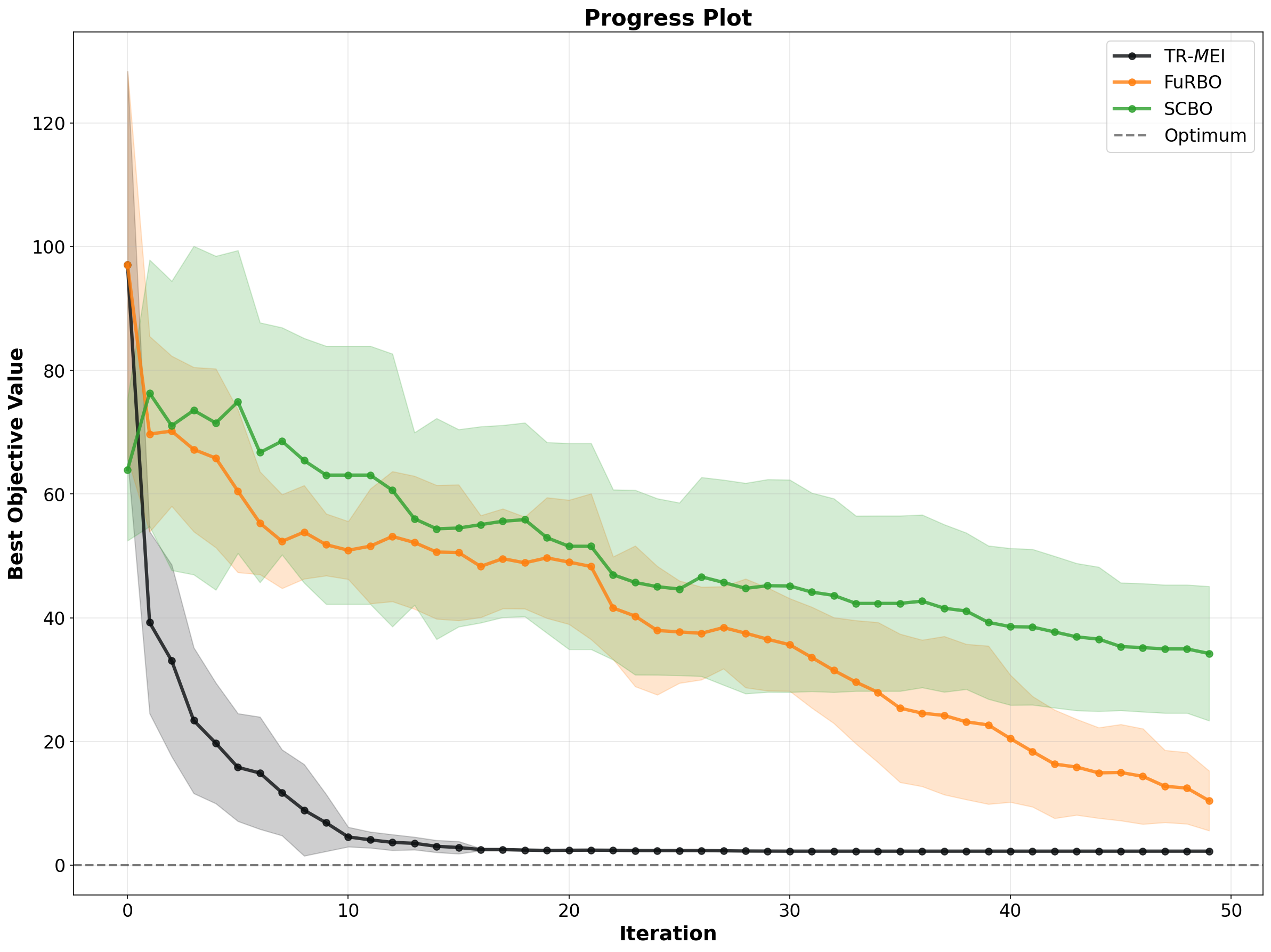}
    \caption{Average objective values obtained over $30$ repeated experiments on L$\hat{e}$vy problem.}
    \label{fig:2}
\end{figure}

\begin{figure}[!ht]
    \centering
    \includegraphics[width=0.95\linewidth]{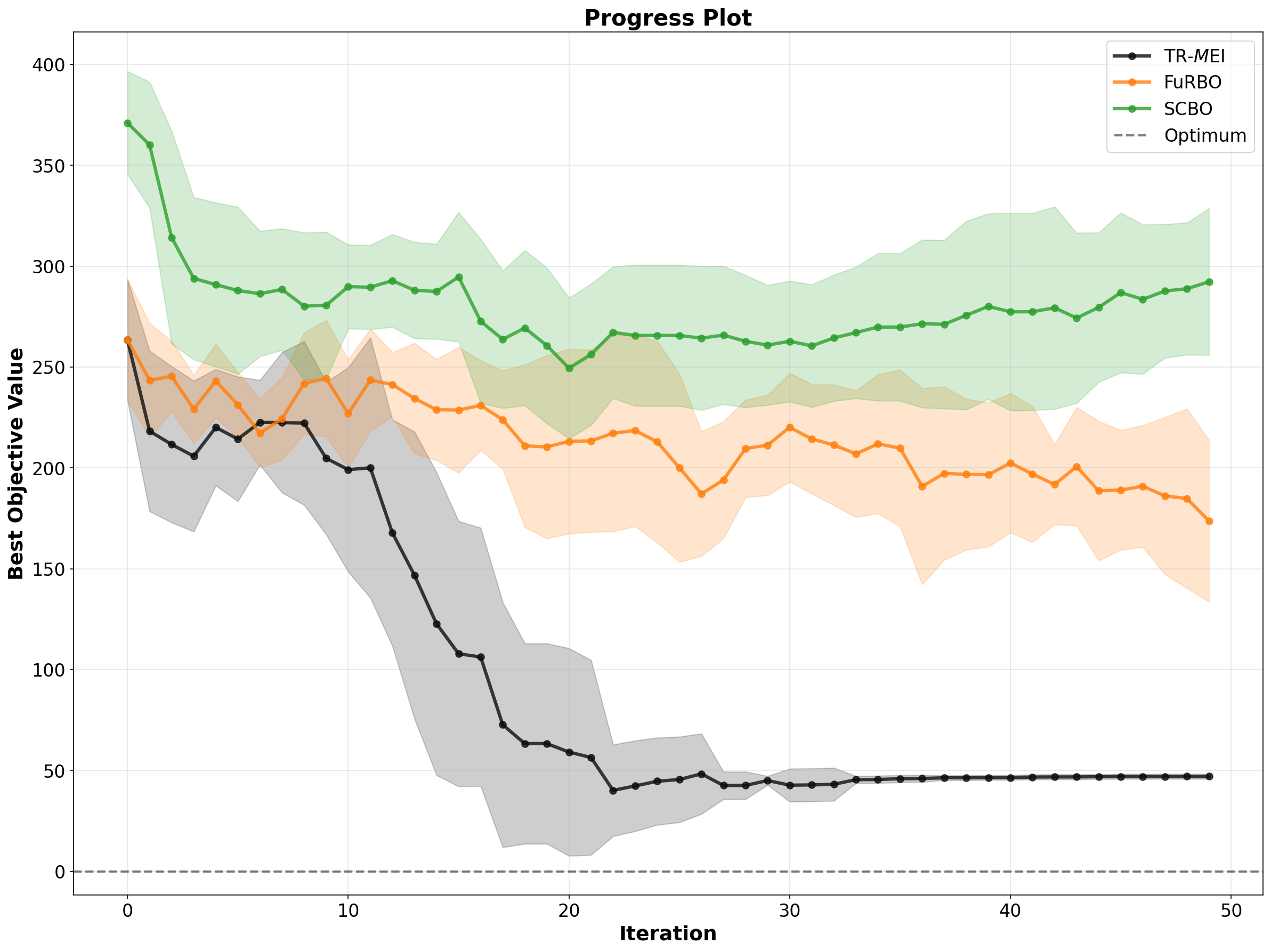}
    \caption{Average objective values obtained over $30$ repeated experiments on Rastrigin problem.}
    \label{fig:3}
\end{figure}

In Figures \ref{fig:1}, \ref{fig:2}, and \ref{fig:3}, we visualize the performance of each method across different benchmark problems. For the Ackley problem, the results show the ability of each method to handle a multimodal landscape and identify feasible solutions. In Figure \ref{fig:1}, TR-\(M\)EI converges to a feasible optimal solution faster than FuRBO and SCBO, which continue to improve over iterations. Table \ref{tab:1} shows that the average performance of TR-\(M\)EI is better than the other methods, which indicates consistent performance.

\begin{table}[!ht]
    \caption{Best valid optimal solution with mean$\pm$standard deviation.}
    \centering
    \begin{tabular}{|c|c|c|c|}
    \hline
    Problems & TR-$M$EI & FuRBO & SCBO \\
    \hline
    Ackley   & $0.0085\pm0.0102$   & $4.4120\pm0.3733$    & $7.3662\pm 1.0541$ \\
    L$\hat{e}$vy   & $0.0071\pm 0.0002$   & $4.8199\pm 0.5856$    & $5.4482\pm 0.3754$ \\
    Rastrigin   & $47.0507\pm 1.1983$   & $173.7190\pm 40.0435$    & $292.2801\pm 36.344$ \\
    \hline
    \end{tabular}
    \label{tab:1}
\end{table}

For the L$\hat{e}$vy problem, the comparison shows how the methods handle strong nonlinearity and complex constraint structures. Figure \ref{fig:2} shows that TR-\(M\)EI reaches a feasible optimal solution around 15 function evaluations, while FuRBO and SCBO approach the optimum only after about 50 evaluations. Table \ref{tab:1} shows the consistency of the solution in terms of average performance.

For the Rastrigin problem, the results show the performance of each method in a setting with many regularly distributed local minima, which makes global optimization difficult. Figure \ref{fig:3} shows that TR-\(M\)EI achieves better performance compared to FuRBO and SCBO. This observation is also supported by Table \ref{tab:1}.

\subsection{Application to portfolio optimization}
Consider a portfolio optimization problem involving \(N\) financial assets discussed in \cite{robert2025}. The objective is to determine an allocation of capital that minimizes financial risk while maintaining a desired level of expected return. Portfolio optimization is widely used in finance because investors seek portfolios that provide stable returns while limiting exposure to market uncertainty.

Let \(\mathbf{Z}=(Z_1,\ldots,Z_N)\) denote the vector of future asset prices. The random vector \(\mathbf{Z}\) is assumed to follow a multivariate normal distribution, where the mean vector and covariance matrix are estimated from historical market data. Let \(\mathbf{x}=(x_1,\ldots,x_N)\) denote the portfolio allocation weights, where \(x_i\) represents the fraction of the available capital invested in the \(i\)-th asset. The portfolio return is defined as
\[
f(\mathbf{x},\mathbf{Z})
=
\sum_{i=1}^{N}x_i\,y_i(Z_i),
\]
where \(y_i(Z_i)\) is the return of the \(i\)-th asset.

To quantify downside risk, we employ the Value-at-Risk (VaR) and Conditional Value-at-Risk (CVaR) measures. For a confidence level \(\alpha\in(0,1)\), the Value-at-Risk is defined as
\[
\mathrm{VaR}_{\alpha}[f(\mathbf{x},\mathbf{Z})] =\inf\left\{\omega:\mathbb{P}\left(f(\mathbf{x},\mathbf{Z})\le-\omega\right)\le1-\alpha\right\},
\]
which represents the largest expected loss that will not be exceeded with probability \(\alpha\).

The Conditional Value-at-Risk is defined as
\[
\mathrm{CVaR}_{\alpha}[f(\mathbf{x},\mathbf{Z})]= -\mathbb{E}\left[f(\mathbf{x},\mathbf{Z})\mid f(\mathbf{x},\mathbf{Z})\le-\mathrm{VaR}_{\alpha}[f(\mathbf{x},\mathbf{Z})]\right].
\]
Unlike VaR, CVaR measures the average loss beyond the VaR threshold and therefore provides a more informative assessment of downside risk.

The portfolio optimization problem seeks an allocation that minimizes the CVaR while satisfying a minimum expected return requirement. The optimization problem is
\begin{align}
\min_{\mathbf{x}}\quad & \mathrm{CVaR}_{\alpha}[f(\mathbf{x},\mathbf{Z})]\\
\text{subject to} \quad & \mathbb{E}_{\mathbf{Z}} \!\left[f(\mathbf{x},\mathbf{Z})\right] \ge r_{\min}, \nonumber \\
& 0\le x_i\le1, \quad i=1,\ldots,N, \nonumber\\
& \sum_{i=1}^{N}x_i\le1.\nonumber
\end{align}

We consider two portfolio optimization problems.

\textbf{Problem 1 (Stock portfolio).} The first problem considers a portfolio consisting of twenty stocks. The return of the \(i\)-th stock is modeled as
\[
y_i(Z_i)=\frac{Z_i}{\bar{Z}_i},
\]
where \(\bar{Z}_i\) is the purchase price of the stock and \(Z_i\) is its future price. The objective is to minimize the portfolio CVaR while maintaining a minimum expected return of \((a)~r_{\min}=1.45, (b)~r_{\min}=1.55\) with \(1-\alpha=0.0001\).

\textbf{Problem 2 (Call option portfolio).} The second problem considers a portfolio of twenty call options written on the same underlying assets. Let \(b_i\) and \(K_i\) denote the current bid price and strike price of the \(i\)-th option. The payoff at maturity is
\[
\max(0,Z_i-K_i),
\]
which leads to the return
\[ y_i(Z_i) = \frac{\max(0,Z_i-K_i)-b_i}{b_i}.
\]
This return captures the nonlinear payoff structure of call options, where the maximum loss is limited to the option premium while the potential gain increases as the underlying asset price rises above the strike price. For this problem, the confidence level is fixed at \(1-\alpha=0.0001\), and the minimum expected return is set to \((a)~r_{\min}=5.30, (b)~r_{\min}=5.40\).

\begin{table}[!ht]
\centering
\caption{Comparison of portfolio performance for different optimization methods.}
\label{tab:2}
\begin{tabular}{|c|c|c|c|c|}
\hline
Problems& & TR-$M$EI & FuRBO & SCBO \\
\hline
Stock&CVaR$(r_{\min} = 1.45)$ & $0.1743 \pm 0.0032$ & $0.1743 \pm 0.0032$ & $0.1935 \pm 0.0020$ \\
portfolio &CVaR$(r_{\min} = 1.55)$ & $0.2405 \pm 0.0027$ & $0.2476 \pm 0.0017$ & $0.2611 \pm 0.0020$ \\
\hline
Call option& CVaR$(r_{\min} = 5.30)$ & $0.2610 \pm 0.0033$ & $0.2777 \pm 0.0017$ & $0.3085 \pm 0.0018$ \\
 portfolio& CVaR$(r_{\min} = 5.40)$ & $0.2970 \pm 0.0017$ & $0.3159 \pm 0.0016$ & $0.3340 \pm 0.0015$ \\
\hline
\end{tabular}
\end{table}
Table \ref{tab:2} summarizes the optimization results obtained by TR-\(M\)EI, SCBO, and FuRBO for both portfolio optimization problems. The reported values correspond to the best feasible solutions obtained by each method under the same experimental setting. Overall, TR-\(M\)EI achieves lower portfolio risk while satisfying the required return constraints, demonstrating its ability to identify high quality feasible portfolios.

\begin{figure}[!ht]
    \centering
    \includegraphics[width=0.48\textwidth]{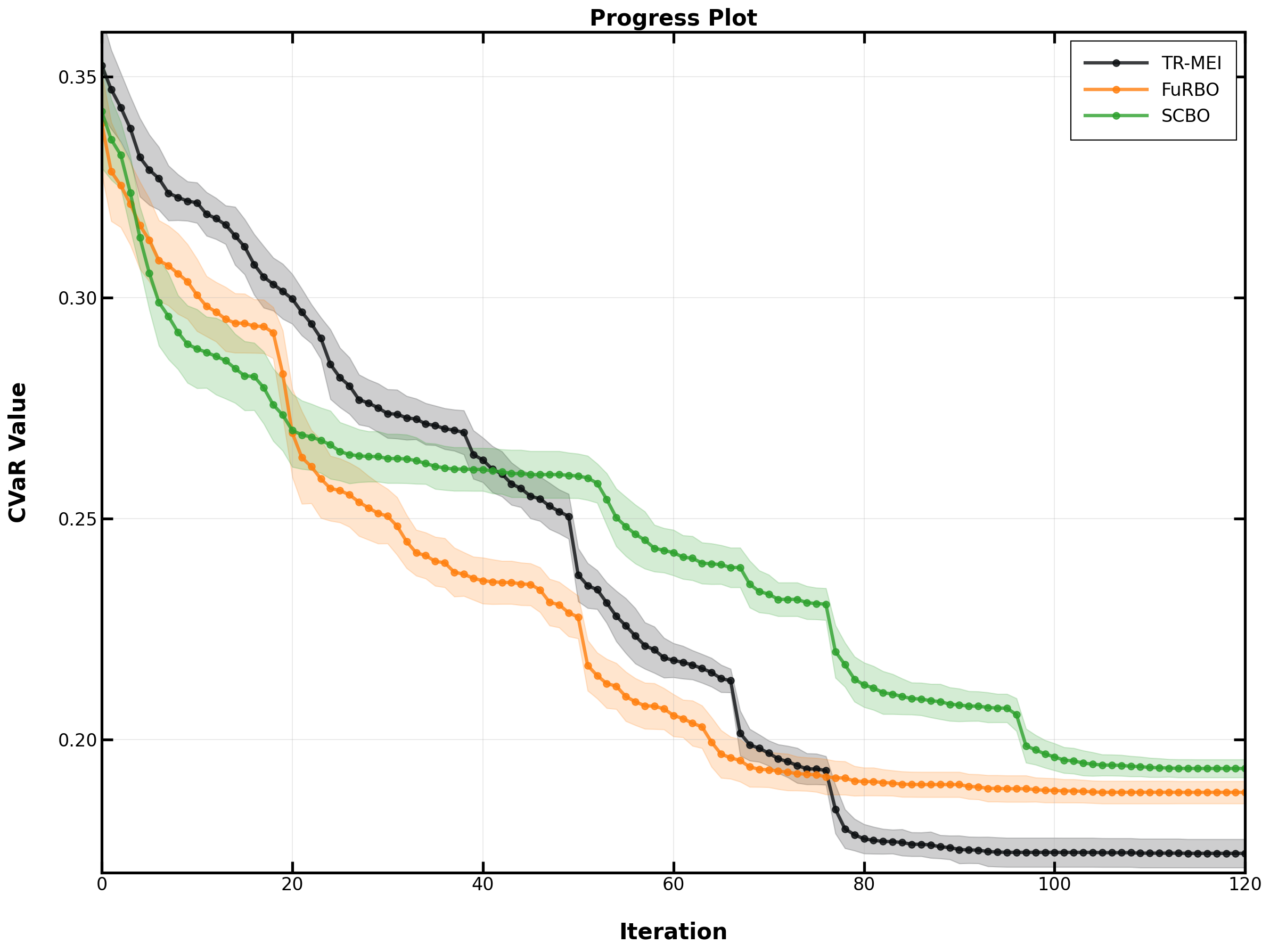}
    \hfill
    \includegraphics[width=0.48\textwidth]{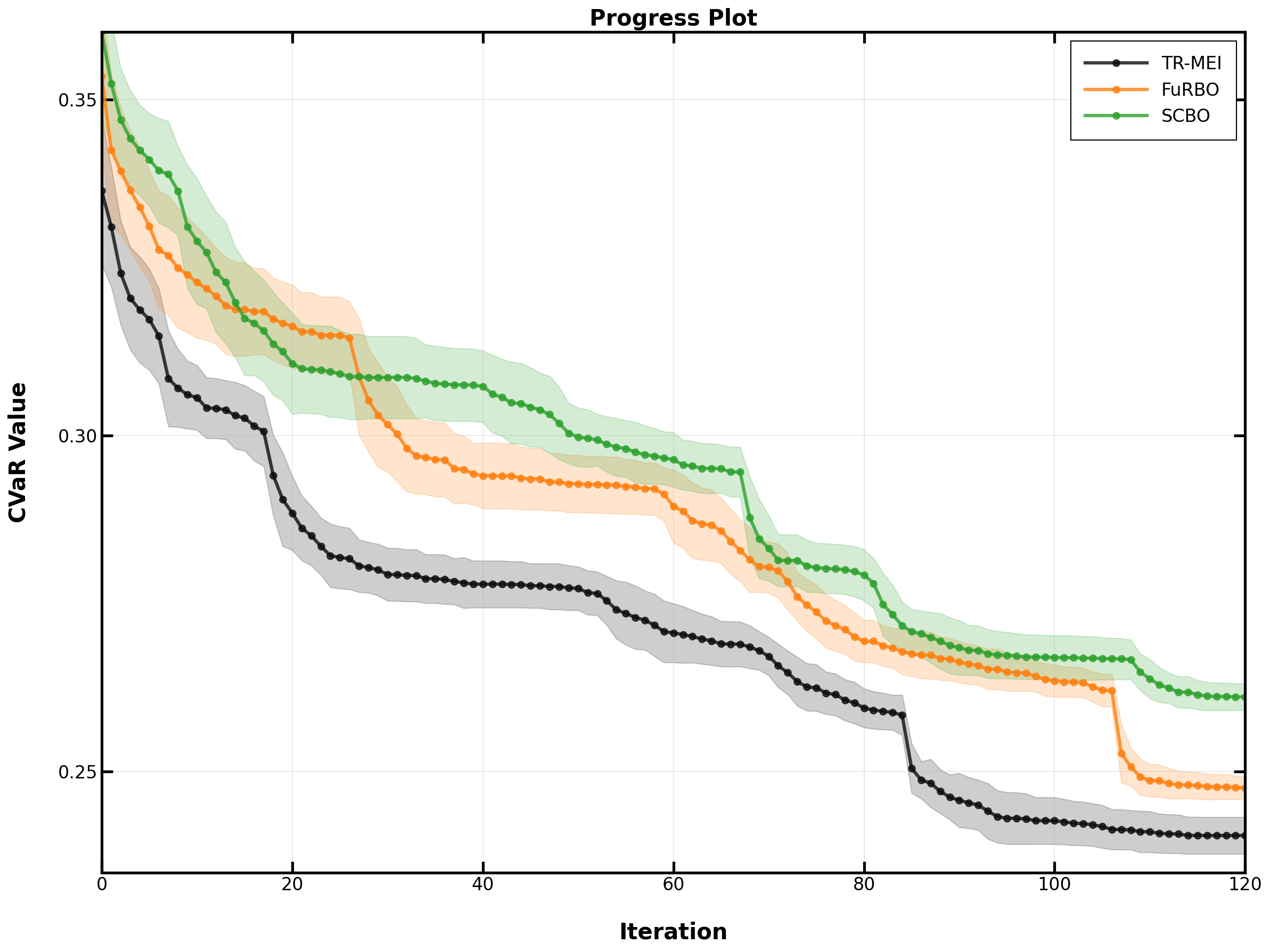}
    \caption{Average objective values obtained over $30$ repeated experiments for the stock portfolio problem with \((a)~r_{\min}=1.45~(\text{left}), (b)~r_{\min}=1.55~(\text{right})\).}
    \label{fig:4}
\end{figure}
\begin{figure}[!ht]
    \centering
    \includegraphics[width=0.48\textwidth]{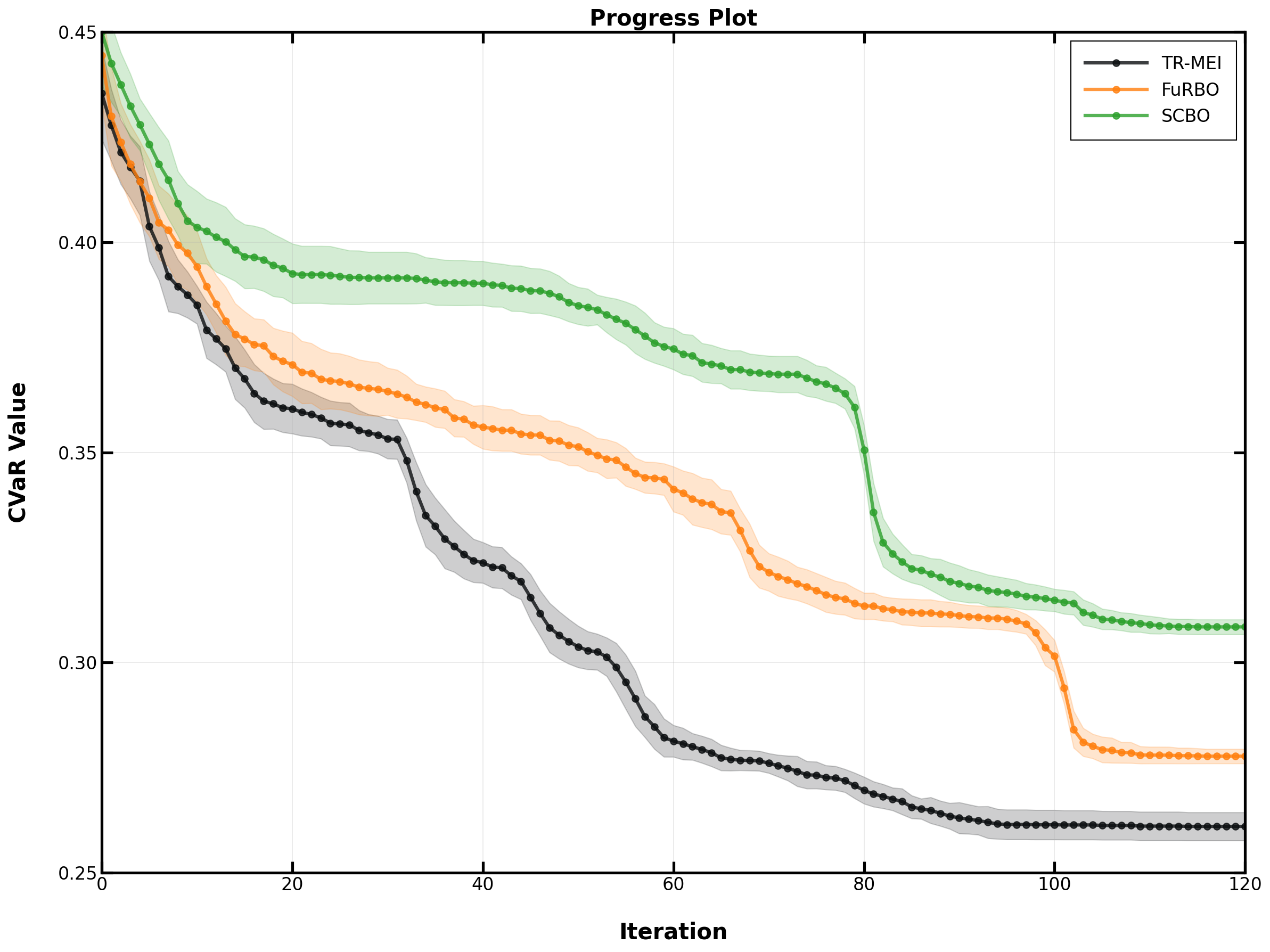}
    \hfill
    \includegraphics[width=0.48\textwidth]{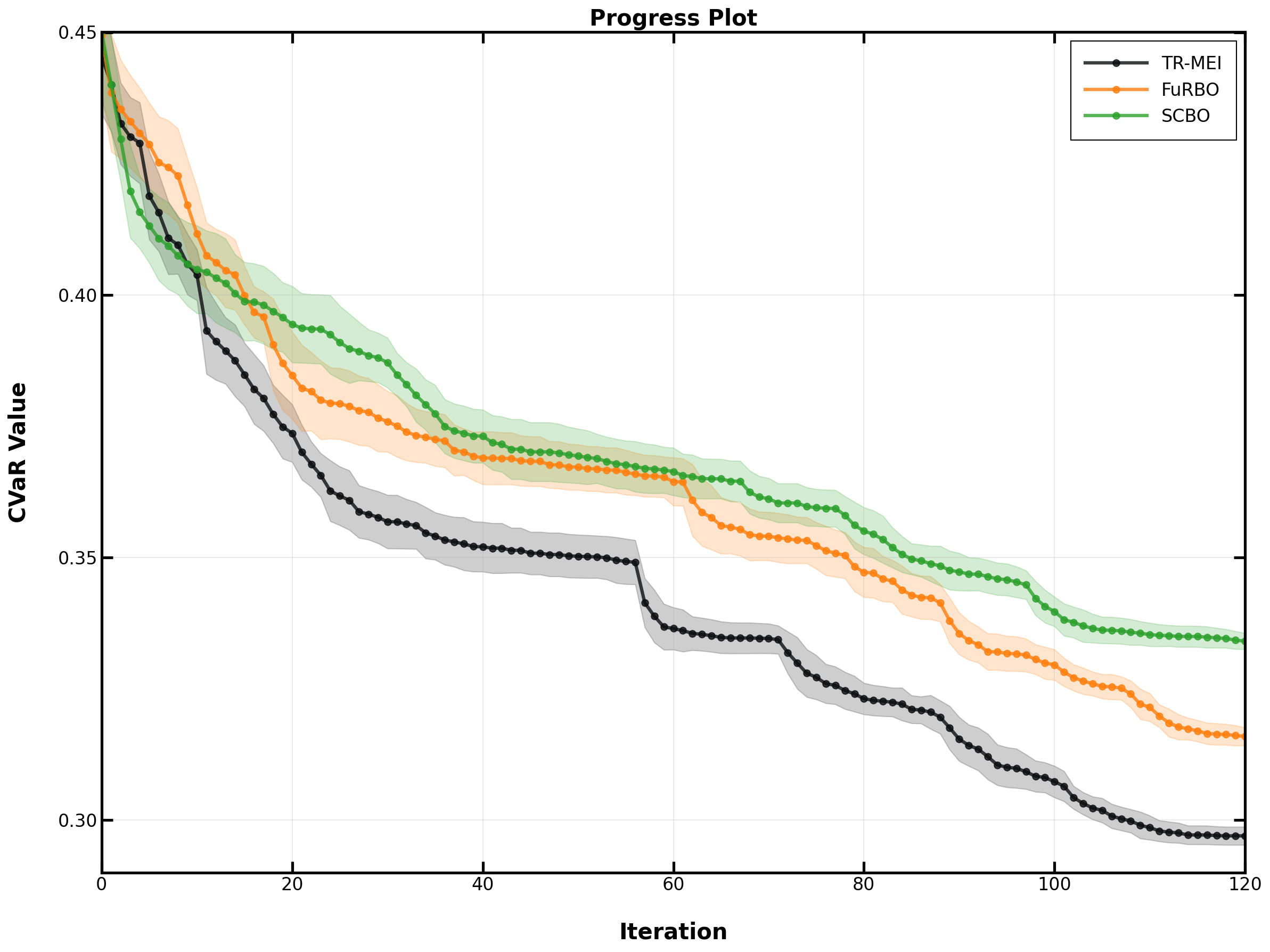}
    \caption{Average objective values obtained over $30$ repeated experiments for the stock portfolio problem with \((a)~r_{\min}=5.30~(\text{left}), (b)~r_{\min}=5.40~(\text{right})\).}
    \label{fig:5}
\end{figure}

Figures \ref{fig:4} and \ref{fig:5} illustrate the convergence behavior of the three methods for the stock portfolio and call option portfolio problems, respectively. Figure \ref{fig:4} shows that TR-\(M\)EI converges more rapidly toward a low-risk feasible portfolio than SCBO and FuRBO. A similar trend is observed in Figure \ref{fig:5}, where TR-\(M\)EI reaches a stable solution with fewer function evaluations, while the competing methods continue to improve over a larger number of evaluations. These results indicate that the proposed trust region strategy, together with the TR-\(M\)EI acquisition function, provides an efficient search mechanism for constrained portfolio optimization.

\section{Discussion}\label{sec5}
An optimization problem in a high-dimensional search space under constraints is challenging for identifying a feasible global optimum. In this article, we propose a hybrid BO methodology that combines the classical big-\(M\) penalty, GP surrogate modeling, and a trust region algorithm. The proposed TR-\(M\)EI method uses the EI acquisition function, which balances exploration and exploitation within a local trust region.

The integration of EI within the trust region enables the method to select informative points based on the predicted mean and uncertainty. As a result, the proposed TR-\(M\)EI approach provides an effective balance between feasibility and optimality while maintaining sample efficiency in high-dimensional constrained optimization problems.

Finally, the comparison of TR-\(M\)EI with SCBO and FuRBO on both synthetic and real-world problems demonstrates its efficiency and effectiveness. The results show that TR-\(M\)EI is able to identify high-quality feasible solutions with fewer function evaluations. It maintains stable performance across different problem settings and scales well with increasing dimensionality. These observations highlight the advantage of combining the penalty-based formulation with a trust region strategy in constrained BO literature.

\appendix

\renewcommand{\thesection}{A.\arabic{section}}
\renewcommand{\theequation}{A\arabic{equation}}
\section*{Appendix A:}

\setcounter{equation}{0}
The mathematical descriptions of the synthetic constrained optimization problems are given below:

\noindent
\textbf{Constrained Ackley Problem}\\
Let \( x = (x_1, \dots, x_d) \in \mathbb{R}^d \). The Ackley function is defined as
\[
f(x) = - a \exp\left(-b \sqrt{\frac{1}{d} \sum_{i=1}^{d} x_i^2}\right) - \exp\left(\frac{1}{d} \sum_{i=1}^{d} \cos(c x_i)\right) + a + e,
\]
where the standard parameter values are
\(
a = 20, \quad b = 0.2, \quad c = 2\pi.
\)

The constrained optimization problem is
\begin{equation}
\label{app_eq1}
\begin{aligned}
\min_{x \in \mathbb{R}^d} \quad & f(x) \\
\text{s.t.} \quad & \sum_{i=1}^{d} x_i \leq 0, \\
& \|x\|_2 \leq 5.
\end{aligned}
\end{equation}

\noindent
\textbf{Constrained L$\hat{e}$vy Problem}\\
Let \( x = (x_1, \dots, x_d) \in \mathbb{R}^d \). Define
\[
w_i = 1 + \frac{x_i - 1}{4}, \quad i = 1, \dots, d.
\]
The optimization problem is
\begin{equation}
\label{app_eq2}
\begin{aligned}
\min_{x \in \mathbb{R}^d} \quad &
\sin^2(\pi w_1) + \sum_{i=1}^{d-1} (w_i - 1)^2 \left[1 + 10 \sin^2(\pi w_i + 1)\right] \\
& \quad + (w_d - 1)^2 \left[1 + \sin^2(2\pi w_d)\right] \\
\text{s.t.} \quad & \sum_{i=1}^{d} x_i \leq 0, \\
& \|x\|_2 \leq 5.
\end{aligned}
\end{equation}

\noindent
\textbf{Constrained Rastrigin Problem}\\
Let \( x = (x_1, \dots, x_d) \in \mathbb{R}^d \). The Rastrigin function is defined as
\[
f(x) = A d + \sum_{i=1}^{d} \left[x_i^2 - A \cos(2\pi x_i)\right],
\]
where the standard parameter value is
\(
A = 10.
\)

The constrained optimization problem is
\begin{equation}
\label{app_eq3}
\begin{aligned}
\min_{x \in \mathbb{R}^d} \quad & f(x) \\
\text{s.t.} \quad & \sum_{i=1}^{d} x_i \leq 0, \\
& \|x\|_2 \leq 5.
\end{aligned}
\end{equation}

\if0\blind{

\section*{Declaration of Interest}
The authors declare no conflicts of interest. All authors contributed equally.
}\fi

\bibliographystyle{apalike}
\spacingset{1}
\bibliography{ref}
	
\end{document}